\documentclass[10pt,twocolumn,letterpaper]{article}

\usepackage{cvpr}
\usepackage{times}
\usepackage{epsfig}
\usepackage{graphicx}
\usepackage{amsmath}
\usepackage{amssymb}

\usepackage{bbm}

\usepackage{subfigure}
\usepackage{caption}

\usepackage{booktabs}  
\usepackage{multirow}

\usepackage[breaklinks=true,bookmarks=false]{hyperref}

\cvprfinalcopy 

\def\cvprPaperID{4483} 

\ifcvprfinal\fi
\begin{document}

\title{AdaCos: Adaptively Scaling Cosine Logits for Effectively
\\ Learning Deep Face Representations}

\author{Xiao Zhang$^{1}$\quad Rui Zhao$^{2}$\quad Yu Qiao$^{3}$\quad Xiaogang Wang$^{1}$\quad Hongsheng Li$^{1}$\\
$^1$CUHK-SenseTime Joint Laboratory, The Chinese University of Hong Kong \quad
$^2$SenseTime Research \\
$^3$SIAT-SenseTime Joint Lab, Shenzhen Institutes of Advanced Technology, Chinese Academy of Sciences\\
{\tt\small zhangx9411@gmail.com\quad \{xgwang, hsli\}@ee.cuhk.edu.hk}
}

\maketitle
\thispagestyle{empty}

\begin{abstract}
The cosine-based softmax losses~\cite{liu_2017_coco_v2, L2-softmax, NormFace,gopal2014mises} and their variants~\cite{CosFace,AM-softmax,ArcFace} achieve great success in deep learning based face recognition. 
However, hyperparameter settings in these losses have significant influences on the optimization path as well as the final recognition performance. Manually tuning those hyperparameters heavily relies on user experience and requires many training tricks.  

In this paper, we investigate in depth the effects of two important hyperparameters of cosine-based softmax losses, the scale parameter and angular margin parameter, by analyzing how they modulate the predicted classification  probability. Based on these analysis, we propose a novel cosine-based softmax loss, AdaCos, which is hyperparameter-free and leverages an adaptive scale parameter to automatically strengthen the training supervisions during the training process. We apply the proposed AdaCos loss to large-scale face verification and identification datasets, including LFW~\cite{LFW}, MegaFace~\cite{MegaFace2}, and IJB-C~\cite{ijbc} 1:1 Verification. Our results show that training deep neural networks with the AdaCos loss is stable and able to achieve high face recognition accuracy. Our method outperforms state-of-the-art softmax losses~\cite{L2-softmax,CosFace,ArcFace} on all the three datasets.
\end{abstract}

\section{Introduction}
Recent years witnessed the breakthrough of deep Convolutional Neural Networks (CNNs) \cite{krizhevsky2012imagenet,hu2017squeeze,parkhi2015deep,szegedy2015going} on significantly improving the performance of  one-to-one $(1:1)$ face verification and one-to-many $(1:N)$ face identification tasks. The successes of deep face CNNs can be mainly credited to three factors: enormous training data \cite{MS-Celeb-1M}, deep neural network architectures \cite{he2016deep,IR} and effective loss functions \cite{L2-softmax,liu_2017_coco_v2,ArcFace}. Modern face datasets, such as LFW~\cite{LFW}, CASIA-WebFace~\cite{WebFace}, MS1M~\cite{MS-Celeb-1M} and MegaFace~\cite{MegaFace1,MegaFace2}, contain huge number of identities which enable the training of deep networks. A number of recent studies, such as DeepFace \cite{DeepFace}, DeepID2~\cite{DeepID2}, DeepID3~\cite{DeepID3}, VGGFace~\cite{parkhi2015deep} and FaceNet~\cite{FaceNet}, demonstrated that properly designed network architectures also lead to improved performance.

Apart from the large-scale training data and deep structures, training losses also play key roles in learning accurate face recognition models \cite{tripletloss2,contrastiveloss,hoffer2015deep}. Unlike image classification tasks, face recognition is essentially an open set recognition problem, where the testing categories (identities) are generally different from those used in training. To handle this challenge, most deep learning based face recognition approaches \cite{DeepID2,DeepID3,DeepFace} utilize CNNs to extract feature representations from facial images, and adopt a metric (usually the cosine distance) to estimate the similarities between pairs of faces during inference.

However, such inference evaluation metric is not well considered in the methods with softmax cross-entropy loss function\footnote{We denote it as ``softmax loss'' for short in the remaining sections.}, which train the networks with the softmax loss but perform inference using cosine-similarities. To mitigate the gap between training and testing, recent works \cite{liu_2017_coco_v2, L2-softmax, NormFace,gopal2014mises} directly optimized cosine-based softmax losses.
Moreover, angular margin-based terms \cite{L-softmax,A-softmax,CosFace,AM-softmax,ArcFace} are usually integrated into cosine-based losses to maximize the angular margins between different identities. 
These methods improve the face recognition performance in the open-set setup. In spite of their successes, the training processes of cosine-based losses (and their variants introducing margins) are usually tricky and unstable. The convergence and performance highly depend on the hyperparameter settings of loss, which are determined empirically through large amount of trials. In addition, subtle changes of these hyperparameters may fail the entire training process.

In this paper, we investigate state-of-the-art cosine-based softmax losses \cite{L2-softmax,CosFace,ArcFace}, especially those aiming at maximizing angular margins, to understand how they provide supervisions for training deep neural networks. Each of the functions generally includes several hyperprameters, which have substantial impact on the final performance and are usually difficult to tune. One has to repeat training with different settings for multiple times to achieve optimal performance. Our analysis shows that different hyperparameters in those cosine-based losses actually have similar effects on controlling the samples' predicted class probabilities. Improper hyperparameter settings cause the loss functions to provide insufficient supervisions for optimizing networks.

Based on the above observation, we propose an adaptive cosine-based loss function, AdaCos, which automatically tunes hyperparameters and generates more effective supervisions during training. 
The proposed AdaCos dynamically scales the cosine similarities between training samples and corresponding class center vectors (the fully-connection vector before softmax), making their predicted class probability meets the semantic meaning of these cosine similarities.
Furthermore, AdaCos can be easily implemented using built-in functions from prevailing deep learning libraries \cite{paszke2017automatic, abadi2016tensorflow,chen2015mxnet,jia2014caffe}. The proposed AdaCos loss leads to faster and more stable convergence for training without introducing additional computational overhead.

To demonstrate the effectiveness of the proposed AdaCos loss function, we evaluated it on several face benchmarks, including LFW face verification \cite{LFW}, MegaFace one-million identification \cite{MegaFace1} and IJB-C \cite{ijbc}. Our method outperforms state-of-the-art cosine-based losses on all these benchmarks.

\section{Related Works}
{\bf Cosine similarities for inference.}
For learning deep face representations, feature-normalized losses are commonly adopted to enhance the recognition accuracy. Coco loss \cite{liu_2017_coco_v1,liu_2017_coco_v2} and NormFace \cite{NormFace} studied the effect of normalization and proposed two strategies by reformulating softmax loss and metric learning. Similarly, Ranjan \etal in \cite{L2-softmax} also discussed this problem and applied normalization on learned feature vectors to restrict them lying on a hypersphere. Movrever, compared with these hard normalization, ring loss~\cite{zheng2018ring} came up with a soft feature normalization approach with convex formulations.

{\bf Margin-based softmax loss.}
Earlier, most face recognition approaches utilized metric-targeted loss functions, such as triplet \cite{tripletloss2} and contrastive loss \cite{contrastiveloss}, which utilize Euclidean distances to measure similarities between features. Taking advantages of these works, center loss \cite{centerloss} and range loss \cite{Zhang_2017_ICCV} were proposed to reduce intra-class variations via minimizing distances within each class \cite{belhumeur1997eigenfaces}. 
Following this, researchers found that constraining margin in Euclidean space is insufficient to achieve optimal generalization. Then angular-margin based loss functions were proposed to tackle the problem. Angular constraints were integrated into the softmax loss function to improve the learned face representation by L-softmax \cite{L-softmax} and A-softmax \cite{A-softmax}. CosFace \cite{CosFace}, AM-softmax \cite{AM-softmax} and ArcFace \cite{ArcFace} directly maximized angular margins and employed simpler and more intuitive loss functions compared with aforementioned methods.

{\bf Automatic hyperparameter tuning.}
The performance of an algorithm highly depends on hyperparameter settings. Grid and random search~\cite{bergstra2012random} are the most widely used strategies. For more automatic tuning, sequential model-based global optimization \cite{hutter2011sequential} is the mainstream choice. Typically, it performs inference with several hyperparameters settings, and chooses setting for the next round of testing based on the inference results. Bayesian optimization \cite{snoek2012practical} and tree-structured parzen estimator approach \cite{bergstra2011algorithms} are two famous sequential model-based methods. However, these algorithms essentially run multiple trials to predict the optimized hyperparameter settings.

\section{Investigation of hyperparameters in cosine-based softmax losses}
In recent years, state-of-the-art cosine-based softmax losses, including L2-softmax~\cite{L2-softmax}, CosFace~\cite{CosFace}, ArcFace~\cite{ArcFace}, significantly improve the performance of deep face recognition. However, the final performances of those losses are substantially affected by their hyperparameters settings, which are generally difficult to tune and require multiple trials in practice. We analyze two most important hyperparameters, the scaling parameter $s$ and the margin parameter $m$, in cosine-based losses. Specially, we deeply study their effects on the prediction probabilities after softmax, which serves as supervision signals for updating entire neural network.

Let $\vec x_{i}$ denote the deep representation (feature) of the $i$-th face image of the current mini-batch with size $N$, and $y_i$ be the corresponding label. The predicted classification probability $P_{i,j}$ of all $N$ samples in the mini-batch can be estimated by the softmax function as
\begin{equation}
P_{i,j}=\frac{e^{f_{i,j}}}{\sum_{k=1}^{C}e^{f_{i,k}}}
\text{, }
\label{prob_for_ij}
\end{equation}
where $f_{i,j}$ is logit used as the input of softmax, $P_{i,j}$ represents its softmax-normalized probability of assigning $\vec x_i$ to class $j$, and $C$ is the number of classes. The cross-entropy loss associated with current mini-batch is
\begin{equation}
\mathcal L_{\text{CE}}=-\frac{1}{N}\sum_{i=1}^{N}\log{P_{i,y_i}}=-\frac{1}{N}\sum_{i=1}^{N}\log{\frac{e^{f_{i,y_i}}}{\sum_{k=1}^{C}e^{f_{i,k}}}}
\text{. }
\end{equation}

Conventional softmax loss and state-of-the-art cosine-based softmax losses~\cite{L2-softmax,CosFace,ArcFace} calculate the logits $f_{i,j}$ in different ways. In conventional softmax loss, logits $f_{i,j}$ are obtained as the inner product between feature $\vec x_{i}$ and the $j$-th class weights $\vec W_{j}$ as $f_{i,j} = {\vec W_{j}^{\text{T}}{\vec x_{i}}} $.
In the cosine-based softmax losses \cite{L2-softmax,CosFace,ArcFace}, cosine similarity is calculated by $\cos \theta_{i,j} = \langle \vec x_{i},\vec W_{j} \rangle / \|\vec{x}_i\| \| \vec{W}_j\|$. The logits $f_{i,j}$ are calculated as $f_{i,j}=s\cdot\cos{\theta_{i,j}}$, where $s$ is a scale hyperparameter. To enforce angular margin on the representations, ArcFace \cite{ArcFace} modified the loss to the form
\begin{equation}
f_{i,j}=s\cdot\cos{(\theta_{i,j}+\mathbbm{1}\{j = y_i\}\cdot{m})}
\text{,}
\label{eq:arcface}
\end{equation}
while CosFace~\cite{CosFace} uses 
\begin{equation}
f_{i,j}=s\cdot(\cos{\theta_{i,j}-\mathbbm{1}\{j = y_i\}\cdot{m})}
\text{,}
\label{eq:cosface}
\end{equation}
where $m$ is the margin. The indicator function $\mathbbm{1}\{ j = y_i\}$ returns $1$ when $j = y_i$ and $0$ otherwise. All margin-based variants decrease $f_{i, y_i}$ associate with the correct class by subtracting  margin $m$. Compared with the losses without margin, margin-based variants require $f_{i, y_i}$ to be greater than other $f_{i,j} \text{ for } j\neq y_i$, by a specified $m$. 

Intuitively, on one hand, the parameter $s$ scales up the narrow range of cosine distances, making the logits more discriminative. On the other hand, the parameter $m$ enlarges the margin between different classes to enhance classification ability. These hyperparameters eventually affect $P_{i,y_i}$. Empirically, an ideal hyperparameter setting should help $P_{i,j}$ to satisfy the following two properties: 
(1) Predicted probabilities $P_{i,y_i}$ of each class (identity) should span to the range $[0,1]$: the lower boundary of $P_{i,y_i}$ should be near $0$ while the upper boundary near $1$; 
(2) Changing curve of $P_{i,y_i}$ should have large absolute gradients around $\theta_{i,y_i}$ to make training effective.

\subsection{Effects of the scale parameter $s$}
\label{ssec:effect_s}

\begin{figure}[t]
\begin{center}
   \includegraphics[width=1\linewidth]{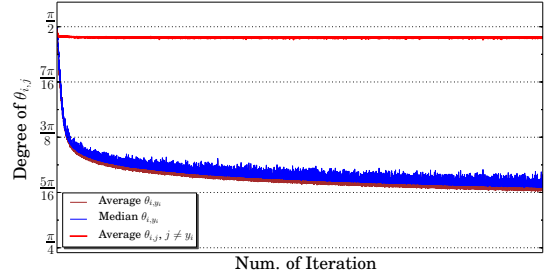}
\end{center}
   \caption{ 
        Changing process of angles in each mini-batch when training on WebFace. (Red) average angles in each mini-batch for non-corresponding classes, $\theta_{i,j}$ for $j \neq y_i$. (Blue) median angles in each mini-batch for corresponding classes, $\theta_{i,y_i}$. (Brown) average angles in each mini-batch for corresponding classes, $\theta_{i,y_i}$.     }
\label{fig:negative_theat_avg}
\end{figure}

The scale parameter $s$ can significantly affect $P_{i,y_i}$. Intuitively, $P_{i,y_i}$ should gradually increase from $0$ to $1$ as the angle $\theta_{i,y_i}$ decreases from $\frac{\pi}{2}$ to $0$\footnote{Mathematically, $\theta$ can be any value in $[0, \pi]$. We empirically found, however, the maximum $\theta$ is always around $\frac{\pi}{2}$. See the red curve in Fig.~\ref{fig:negative_theat_avg} for examples.}, \ie, the smaller the angle between $\vec{x}_i$ and its corresponding class weight $\vec{W}_{y_i}$ is, the larger the probability should be. Both improper probability range and probability curves w.r.t. $\theta_{i,y_i}$ would negatively affect the training process and thus the recognition performance.

We first study the range of classification probability $P_{i,j}$. Given scale parameter $s$, the range of probabilities in all cosine-based softmax losses is
\begin{equation}
\frac{1}{1+(C-1)\cdot e^{s}}\leq{P_{i,j}}\leq{\frac{e^{s}}{e^{s}+(C-1)}},
\end{equation}
where the lower boundary is achieved when $f_{i,j} = s \cdot 0 = 0$ and $f_{i,k} = s\cdot 1 = s$ for all $k \neq j$ in Eq.~\eqref{prob_for_ij}. Similarly, the upper bound is achieved when $f_{i,j} = s$ and $f_{i,k} = 0$ for all $k \neq j$. The range of $P_{i,j}$ approaches 1 when $s\rightarrow \infty$, \ie,
\begin{equation}
\lim\limits_{s \to +\infty }\left(\frac{e^{s}}{e^{s}+(C-1)}-\frac{1}{1+(C-1)\cdot e^{s}}\right)=1,
\end{equation}
which means that the requirement of the range spanning $[0,1]$ could be satisfied with a large $s$. However it does not mean that the larger the scale parameter, the better the selection is. In fact the probability range can easily approach a high value, such as ${0.94}$ when class number $C=10$ and scale parameter $s=5.0$. But an oversized scale would lead to poor probability distribution, as will be discussed in the following paragraphs.

\begin{figure*}
  \begin{center}
  \subfigure[ $P_{i,y_i}$  w.r.t. $\theta_{i,y_i}$.]{
  \begin{minipage}[t]{1\linewidth}
  \centering
  \includegraphics[width=18cm]{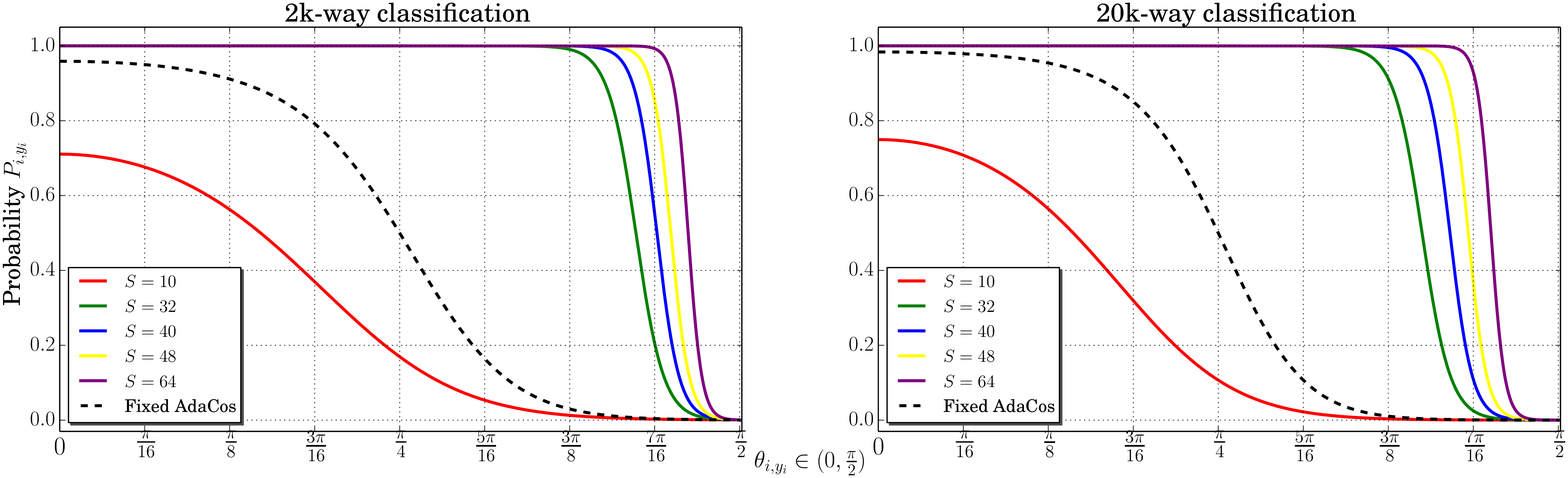}
  \label{fig:theta_prob_mapping_for_diff_scale}
  \end{minipage}
  }
  
  \subfigure[ $P_{i,y_i}$  w.r.t. $\theta_{i,y_i}$.]{
  \begin{minipage}[t]{1\linewidth}
  \centering
  \includegraphics[width=18cm]{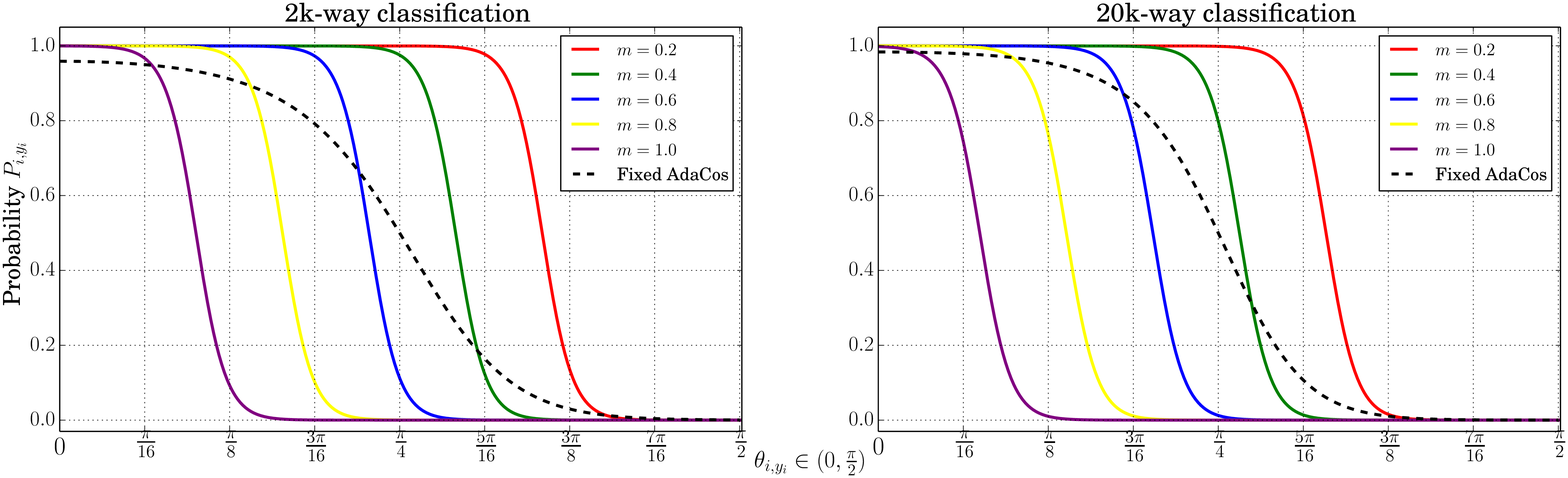}
  \label{fig:theta_prob_mapping_for_diff_margin}
  \end{minipage}
  }
  \end{center}
\caption{ Curves of $P_{i,y_i}$  w.r.t. $\theta_{i,y_i}$ by choosing different sclae and margin parameters. (Left) $C=2000$. (Right) $C=20000$. Fig.~\ref{fig:theta_prob_mapping_for_diff_scale} is for choosing different scale parameters and Fig.~\ref{fig:theta_prob_mapping_for_diff_margin} is for fixing $s=30$ and choosing different margin parameters.}
\label{fig:2d_features_on_s}
\end{figure*}

We investigate the influences of parameter $s$ by taking $P_{i, y_i}$ as a function of $s$ and angle $\theta_{i,y_i}$ where $y_i$ denotes the label of $\vec x_i$. Formally, we have 
\begin{equation}
P_{i,y_i}=\frac{e^{f_{i,y_i}}}{e^{f_{i,y_i}}+B_{i}}=\frac{e^{s\cdot\cos{\theta_{i,y_i}}}}{e^{s\cdot\cos{\theta_{i,y_i}}}+B_{i}},
\label{prob_for_ij_with_B}
\end{equation}
where $B_i =\sum_{k\neq y_i} e^{f_{i,k}} = \sum_{k\neq y_i} e^{s\cdot \cos\theta_{i,k}}$ are the logits summation of all non-corresponding classes for feature $\vec{x}_i$. We observe that the values of $B_i$ are almost unchanged during the training process. This is because the angles $\theta_{i,k}$ for non-corresponding classes $k\neq y_i$ always stay around $\frac{\pi}{2}$ during training (see red curve in Fig. \ref{fig:negative_theat_avg}). 

Therefore, we can assume $B_i$ is constant, \ie, $B_i \approx \sum_{k\neq y_i} e^{s\cdot \cos (\pi/2)}$ $ = C-1$. We then plot curves of probabilities $P_{i,y_i}$ w.r.t. $\theta_{i, y_i}$ under different setting of parameter $s$ in Fig. \ref{fig:theta_prob_mapping_for_diff_scale}. It is obvious that when $s$ is too small (\eg, $s=10$ for class/identity number $C=2,000$ and $C=20,000$), the maximal value of $P_{i,y_i}$ could not reach $1$. This is undesirable because even when the network is very confident on a sample $\vec{x_i}$'s corresponding class label $y_i$, e.g. $\theta_{i,y_i}=0$, the loss function would still penalize the classification results and update the network. 

On the other hand, when $s$ is too large (\eg, $s=64$), the probability curve $P_{i,y_i}$ w.r.t. $\theta_{i,y_i}$ is also problematic. It would output a very high probability even when $\theta_{i,y_i}$ is close to $\pi/2$, which means that the loss function with large $s$ may fail to penalize mis-classified samples and cannot effectively update the networks to correct mistakes.

In summary, the scaling parameter $s$ has substantial influences to the range as well as the curves of the probabilities $P_{i,y_i}$, which are crucial for effectively training the deep network.

\subsection{Effects of the margin parameter $m$}
\label{ssec:effect_m}
In this section, we investigate the effect of margin parameters $m$ in cosine-based softmax losses (Eqs. \eqref{eq:arcface} \& \eqref{eq:cosface}), and their effects on feature $\vec{x}_i$'s predicted class probability $P_{i,y_i}$. For simplicity, we here study the margin parameter $m$ for ArcFace (Eq. \ref{eq:arcface}); while the similar conclusions also apply to the parameter $m$ in CosFace (Eq. \eqref{eq:cosface}).

We first re-write classification probability $P_{i,y_i}$ following Eq. \eqref{prob_for_ij_with_B} as
\begin{equation}
P_{i,y_i}=\frac{e^{f_{i,y_i}}}{e^{f_{i,y_i}}+B_{i}}=\frac{e^{s\cdot\cos{(\theta_{i,y_i}+m)}}}{e^{s\cdot\cos{(\theta_{i,y_i}+m)}}+B_{i}}
\text{. }
\label{prob_for_ij_with_B_margin}
\end{equation}
To study the influence of parameter $m$ on the probability $P_{i,y_i}$, we assume both $s$ and $B_i$ are fixed. Following the discussion in Section \ref{ssec:effect_s}, we set $B_i\approx{C-1}$, and fix $s=30$. The probability curves $P_{i,y_i}$ w.r.t. $\theta_{i,y_i}$ under different $m$ are shown in Fig.~\ref{fig:theta_prob_mapping_for_diff_margin}.

According to Fig.~\ref{fig:theta_prob_mapping_for_diff_margin}, increasing the margin parameter shifts probability $P_{i,y_i}$ curves to the left. Thus, with the same $\theta_{i,y_i}$, larger margin parameters lead to lower probabilities $P_{i,y_i}$ and thus larger loss even with small angles $\theta_{i,y_i}$. In other words, the angles $\theta_{i,y_i}$ between the feature $\vec{x}_i$ and its corresponding class's weights $\vec{W}_{y_i}$ have to be very small for sample $i$ being correctly classified. This is the reason why margin-based losses provide stronger supervisions for the same $\theta_{i,y_i}$ than conventional cosine-based losses. Proper margin settings have shown to boost the final recognition performance in \cite{CosFace,ArcFace}.

Although larger margin $m$ provides stronger supervisions, it should not be too large either. When $m$ is oversized (\eg, $m=1.0$), the probabilities $P_{i,y_i}$ becomes unreliable. It would output probabilities around $0$ even $\theta_{i,y_i}$ is very small. This lead to large loss for almost all samples even with very small  sample-to-class angles, which makes the training difficult to converge. In previous methods, the margin parameter selection is an ad-hoc procedure and has no theoretical guidance for most cases.

\subsection{Summary of the hyparameter study}
According to our analysis, we can draw the following conclusions:

(1) Hyperparameters scale $s$ and margin $m$ can substantially influence the prediction probability $P_{i,y_i}$ of feature $\vec{x}_i$ with ground-truth identity/category $y_i$. 
For the scale parameter $s$, too small $s$ would limit the maximal value of $P_{i,y_i}$. On the other hand, too large $s$ would make most predicted probabilities $P_{i,y_i}$ to be $1$, which makes the training loss insensitive to the correctness of $\theta_{i,y_i}$.
For the margin parameter $m$, a too small margin is not strong enough to regularize the final angular margin, while an oversized margin makes the training difficult to converge.

(2) The effect of scale $s$ and margin $m$ can be unified to modulate the mapping from cosine distances $\cos{\theta_{i,y_i}}$ to the prediction probability $P_{i,y_i}$. As shown in Fig.~\ref{fig:theta_prob_mapping_for_diff_scale} and Fig.~\ref{fig:theta_prob_mapping_for_diff_margin}, both small scales and large margins have similar effect on $\theta_{i,y_i}$ for strengthening the supervisions, while both large scales and small margins weaken the supervisions. Therefore it is feasible and promising to control the probability $P_{i,y_i}$  using one single hyperparameter, either $s$ or $m$. Considering the fact that $s$ is more related to the range of $P_{i,y_i}$ that required to span $[0,1]$, we will focus on automatically tuning the scale parameter $s$ in the reminder of this paper.

\section{The cosine-based softmax loss with adaptive scaling}
Based on our previous studies on the hyperparameters of the cosine-based softmax loss functions, in this section, we propose a novel loss with a self-adaptive scaling scheme, namely AdaCos, which does not require the ad-hoc and time-consuming manual parameter tuning. Training with the proposed loss does not only facilitate convergence but also results in higher recognition accuracy.

Our previous studies on Fig. \ref{fig:negative_theat_avg} show that during the training process, the angles $\theta_{i,k}$ for $k\neq y_i$ between the feature $\vec{x}_i$ and its non-corresponding weights $\vec{W}_{k\neq y_i}$ are almost always close to $\frac{\pi}{2}$, In other words, we could safely assume that $B_i \approx \sum_{k\neq y_i} e^{s\cdot \cos(\pi/2)} = C-1$ in Eq. \eqref{prob_for_ij_with_B}. Obviously, it is the probability $P_{i,y_i}$ of feature $x_i$ belonging to its corresponding class $y_i$ that has the most influence on supervision for network training. Therefore, we focus on designing an adaptive scale parameter for controling the probabilities $P_{i,y_i}$.

From the curves of $P_{i,y_i}$ w.r.t. $\theta_{i,y_i}$ (Fig.~\ref{fig:theta_prob_mapping_for_diff_scale}), we observe that the scale parameter $s$ does not only simply affect $P_{i,y_i}$'s boundary of of determining correct/incorrect but also squeezes/stretches the $P_{i,y_i}$ curvature; In contrast to scale $s$, margin parameter $m$ only shifts the curve in phase. 
We therefore propose to automatically tune the scale parameter $s$ and eliminate the margin parameter $m$ from our loss function, which makes our proposed AdaCos loss different from state-of-the-art softmax loss variants with angular margin. With softmax function, the predicted probability can be defined by
\begin{align}
	P_{i,j} = \frac{e^{\tilde{s}\cdot \cos \theta_{i,j}}}{\sum_{k=1}^C e^{\tilde{s} \cdot \cos \theta_{i,k}}},
\end{align}
where $\tilde{s}$ is the automatically tuned scale parameter to be discussed below.

Let us first re-consider the $P_{i,y_i}$ (Eq. \eqref{prob_for_ij_with_B}) 
as a function of $\theta_{i,y_i}$. Note that $\theta_{i,y_i}$ represents the angle between sample $\vec x_i$ and the weight vector of its ground truth category $y_i$. For network training, we hope to minimize $\theta_{i,y_i}$ with the supervision from the loss function $L_{\text{CE}}$. Our objective is choose a suitable scale $\tilde{s}$ which makes predicted probability $P_{i,y_i}$ change significantly with respect to $\theta_{i,y_i}$. Mathematically, we find the point where the absolute gradient value  $\| \frac{ \partial P_{i,y_i}(\theta)}{ \partial \theta }\| $ reaches its maximum, when the second-order derivative of $P_{i,y_i}$ at $\theta_{0}$ equals $0$, \ie,
\begin{equation}
\frac{\partial^2 P_{i,y_i}(\theta_{0})}{\partial {\theta_{0}}^2}=0
\text{, }
\label{inflection_point_second_order}
\end{equation}
where $\theta_{0}\in [0,\frac{\pi}{2}]$. Combining Eqs.~\eqref{prob_for_ij_with_B} and~\eqref{inflection_point_second_order}, we obtain an transcendental equation. Considering that $P(\theta_0)$ is close to $\frac{1}{2}$, the relation between the scale parameter $s$ and the point $(\theta_{0},P(\theta_{0}))$ can be approximated as
\begin{equation}
s_{0}=\frac{\log{B_i}}{\cos{\theta_{0}}}
\text{, }
\label{s_vs_theta0}
\end{equation}
where $B_i$ can be well approximated as $B_i= \sum_{k\neq y_i} e^{s\cdot \cos\theta_{i,k}} \approx C-1$ since the angles $\theta_{i,k}$ distribute around $\pi/2$ during training (see Eq.~\eqref{prob_for_ij_with_B} and Fig.~\ref{fig:negative_theat_avg}). Then the task of automatically determining $\tilde{s}$ would reduce to select an reasonable central angle $\tilde{\theta}$ in $[0,\pi/2]$.

\subsection{Automatically choosing a fixed scale parameter}
Since $\frac{\pi}{4}$ is in the center of $[0,\frac{\pi}{2}]$, it is natural to regard $\pi/4$ as the point, \ie setting $\theta_0=\pi/4$ for figuring out an effective mapping from angle $\theta_{i, y_i}$ to the probability $P_{i, y_i}$.
Then the supervisions determined by $P_{i,y_i}$ would be back-propagated to update $\theta_{i,y_i}$ and further to update network parameters. 
According to Eq. \eqref{s_vs_theta0}, we can estimate the corresponding scale parameter $s_f$ as
\begin{align}
	\tilde{s}_f = \frac{\log B_i}{\cos {\frac{\pi}{4}}} &= \frac{\log \sum_{k\neq y_i}e^{s\cdot \cos\theta_{i,k}}}{\cos \frac{\pi}{4}} \label{naive_s} \\ &\approx \sqrt{2}\cdot\log{(C-1)} \nonumber
\end{align}
where $B_i$ is approximated by $C-1$.

For such an automatically-chosen fixed scale parameter $\tilde{s}_f$ (see Figs. \ref{fig:theta_prob_mapping_for_diff_scale} and \ref{fig:theta_prob_mapping_for_diff_margin}), it depends on the number of classes $C$ in the training set and also provides a good guideline for existing cosine distance based softmax losses to choose their scale parameters. In contrast, the scaling parameters in existing methods was manually set according to human experience. It acts as a good baseline method for our dynamically tuned scale parameter $\tilde{s}_d$ in the next section.

\subsection{Dynamically adaptive scale parameter}
As Fig. \ref{fig:negative_theat_avg} shows, the angles $\theta_{i,y_i}$ between features $\vec{x}_i$ and their ground-truth class weights $\vec{W}_{y_i}$ gradually decrease as the training iterations increase; while the angles between features $\vec{x}_i$ and non-corresponding classes $\vec{W}_{j\neq y_i}$ become stabilize around $\frac{\pi}{2}$, as shown in Fig.~\ref{fig:negative_theat_avg}.

Although our previously fixed scale parameter $\tilde{s}_f$ behaves properly as $\theta_{i, y_i}$ changes over $[0,\frac{\pi}{2}]$, it does not take into account the fact that $\theta_{i,y_i}$ gradually decrease during training. Since smaller $\theta_{i,y_i}$ gains higher probability $P_{i,y_i}$ and thus gradually receives weaker supervisions as the training proceeds, we therefore propose a dynamically adaptive scale parameter $\tilde{s}_d$ to gradually apply stricter requirement on the position of $\theta_0$ which can progressively enhance the supervisions throughout the training process.

Formally we introduce a modulating indicator variable $\theta^{(t)}_{\text{med}}$, which is the median of all corresponding classes' angles, $\theta_{i,y_i}^{(t)}$, from the mini-batch of size $N$ at the $t$-th iteration. $\theta^{(t)}_{\text{med}}$ roughly represents the current network's degree of optimization on the mini-batch. When the median angle is large, it denotes that the network parameters are far from optimum and less strict supervisions should be applied to make the training converge more stably;  when the median angle $\theta_{\text{med}}^{(t)}$ is small, it denotes that the network is close to optimum and stricter supervisions should be applied to make the intra-class angles $\theta_{i,y_i}$ become even smaller. Based on this observation, we set the central angle $\tilde{\theta}^{(t)}_{0} = \theta^{(t)}_{\text{med}}$. We also introduce $B_{\text{avg}}^{(t)}$ as the average of $B_i^{(t)}$ as 
\begin{align}
	B_\text{avg}^{(t)}= \frac{1}{N} \sum_{i \in \mathcal{N}^{(t)}} B_i^{(t)} = \frac{1}{N} \sum_{i \in \mathcal{N}^{(t)}} \sum_{k\neq y_i} e^{\tilde{s}_d^{(t-1)} \cdot \cos \theta_{i,k}} ,
\end{align}
where $\mathcal{N}^{(t)}$ denotes the face identity indices in the mini-batch at the $t$-th iteration. Unlike approximating $B_i \approx C-1$ for the fixed adaptive scale parameter $\tilde{s}_f$, here we estimate $B_i^{(t)}$ using the scale parameter $\tilde{s}_d^{(t-1)}$ of previous iteration, which provides us a more accurate approximation. 
Be reminded that $B_i^{(t)}$ also includes dynamic scale  $\tilde{s}_d^{(t)}$. We can obtain it by solving the nonlinear function given by the above equation. In practice, we notice that $\tilde{s}_d^{(t)}$ changes very little following iterations. So, we just use $\tilde{s}_d^{(t-1)}$ to calculate $B_i^{(t)}$  with Eq.~\eqref{prob_for_ij_with_B}. Then we can obtain dynamic scale $\tilde{s}_d^{(t)}$ directly with Eq.~\eqref{s_vs_theta0}. So we have:
\begin{align}
	\tilde{s}_{d}^{(t)} = \frac{\log B_\text{avg}^{(t)}}{\cos \theta_\text{med}^{(t)}},
\end{align}
where $B_\text{avg}^{(t)}$ is related to the dynamic scale parameter. We estimate it using the scale parameter $\tilde{s}_d^{(t-1)}$ of the previous iteration.

At the begin of the training process, the median angle $\theta_\text{med}^{(t)}$ of each mini-batch might be too large to impose enough supervisions for training. We therefore force the central angle $\theta_{\text{med}}^{(t)}$ to be less than $\frac{\pi}{4}$.
Our dynamic scale parameter for the $t$-th iteration could then be formulated as
\begin{align}
	\label{AdaCos}
	\tilde{s}_d^{(t)} = 
	\begin{cases}
		\sqrt{2}\cdot\log{(C-1)} & t = 0,\\
		\displaystyle \frac{\log{B_\text{avg}^{(t)}}}{\cos\left(\min(\frac{\pi}{4},\theta_\text{med}^{(t)})\right)} & t \geq 1,
	\end{cases}
\end{align}
where $\tilde{s}_d^{(0)}$ is initialized as our fixed scale parameter $\tilde{s}_f$ when $t=0$.

Substituting $\tilde{s}_d^{(t)}$ into $f_{i,j}=\tilde{s}_d^{(t)}\cdot\cos{\theta_{i,j}}$, the corresponding gradients can be calculated as follows
\begin{equation}
\begin{aligned}
&\frac{\partial\mathcal{L}(\Vec{x}_i)}{\partial \Vec{x}_i}=\sum^{C}_{j=1}(P_{i,j}^{(t)} -\mathbbm{1}( y_i =j))\cdot{\tilde{s}_d^{(t)}}\frac{\partial \cos{\theta_{i,j}}}{\partial {\Vec{x}_i}}
\text{, }
\\
&\frac{\partial\mathcal{L}(\Vec{W}_j)}{\partial \Vec{W}_j}=(P_{i,j}^{(t)} - \mathbbm{1}( y_i =j))\cdot{\tilde{s}_d^{(t)}}\frac{\partial \cos{\theta_{i,j}}}{\partial {\Vec{W}_j}}
\text{, }
\label{gradient_f_w}
\end{aligned}
\end{equation}
where $\mathbbm{1}$ is the indicator function and
\begin{equation}
P_{i,j}^{(t)}=\frac{e^{\tilde{s}_d^{(t)}\cdot\cos{\theta_{i,j}}}}{\sum_{k=1}^{C}e^{\tilde{s}_d^{(t)}\cdot\cos{\theta_{i,k}}}}\text{. }
\label{p_ij_T}
\end{equation}
Eq.~\eqref{p_ij_T} shows that the dynamically adaptive scale parameter $\tilde{s}_d^{(t)}$ influences classification probabilities differently at each iteration and also effectively affects the gradients (Eq.~\eqref{gradient_f_w}) for updating network parameters. The benefit of dynamic AdaCos is that it can produce reasonable scale parameter by sensing the training convergence of the model in the current iteration.

\section{Experiments}
We examine the proposed AdaCos loss function on several public face recognition benchmarks and compare it with state-of-the-art cosine-based softmax losses. The compared losses include $l2$-softmax~\cite{L2-softmax}, CosFace~\cite{CosFace}, and ArcFace~\cite{ArcFace}. We present evaluation results on LFW~\cite{LFW}, MegaFace 1-million Challenge~\cite{MegaFace1}, and IJB-C~\cite{ijbc} data. We also present results on some exploratory experiments to show the convergence speed and robustness against low-resolution images.

\textbf{Preprocessing.} We use two public training datasets, CASIA-WebFace~\cite{WebFace} and MS1M~\cite{MS-Celeb-1M}, to train CNN models with our proposed loss functions. We carefully clean the noisy and low-quality images from the datasets. The cleaned WebFace~\cite{WebFace} and MS1M~\cite{MS-Celeb-1M} contain about $0.45$M and $2.35$M facial images, respectively. All models are trained based on these training data and directly tested on the test splits of the three datasets. RSA~\cite{liu_2017_rsa} is applied to the images to extract facial areas. Then, according to detected facial landmarks, the faces are aligned through similarity transformation and resized to the size $144\times 144$. All image pixel values are subtracted with the mean $127.5$ and dividing by $128$.

\subsection{Results on LFW}
The LFW~\cite{LFW} dataset collected thousands of identities from the inertnet. Its testing protocol contains about $13,000$ images for about $1,680$ identities with a total of $6,000$ ground-truth matches. Half of the matches are positive while the other half are negative ones. LFW's primary difficulties lie in face pose variations, color jittering, illumination variations and aging of persons. Note portion of the pose variations can be eliminated by the RSA~\cite{liu_2017_rsa} facial landmark detection and alignment algorithm, but there still exist some non-frontal facial images which can not be aligned by RSA~\cite{liu_2017_rsa} and then aligned manually.

\subsubsection{Comparison on LFW}
For all experiments on LFW~\cite{LFW}, we train ResNet-50 models~\cite{he2016deep} with batch size of $512$ on the cleaned WebFace~\cite{WebFace} dataset. The input size of facial image is $144\times144$ and the feature dimension input into the loss function is $512$. Different loss functions are compared with our proposed AdaCos losses.

\begin{table}
\begin{center}
\begin{tabular}{|l|c|c|c|c|}
\hline
Method & $1$st & $2$nd & $3$rd& Average Acc. \\
\hline\hline
Softmax      & $93.05$ &$92.92$ &$93.27$ & $93.08$\\
$l2$-softmax~\cite{L2-softmax} & $98.22$ &$98.27$ &$98.08$ & $98.19$\\
CosFace~\cite{CosFace}      & $99.37$ &$99.35$ &$99.42$ & $99.38$\\
ArcFace~\cite{ArcFace}      & $99.55$ &$99.37$ &$99.43$ & $99.45$\\
\hline
Fixed AdaCos& $99.63$ &$99.62$ &$99.55$ & $\textbf{99.60}$\\
\bf{Dyna. AdaCos}     & $99.73$ &$99.72$ &$99.68$ & $\textbf{99.71}$\\
\hline
\end{tabular}
\end{center}
\caption{ Recognition accuracy on LFW by ResNet-50 trained with different compared losses. All the methods are trained on the cleaned WebFace~\cite{WebFace} training data and tested on LFW for three times to obtain the average accuracy.}
\label{tab:lfw_benchmark}

\end{table}

\begin{figure}[t]
\begin{center}
   \includegraphics[width=1\linewidth]{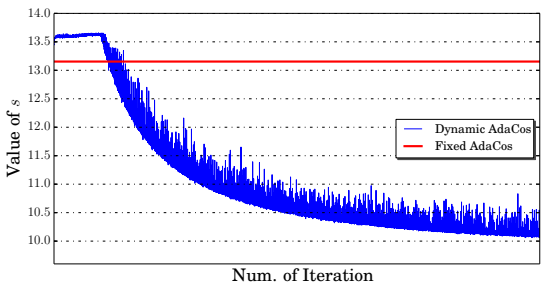}
\end{center}
   \caption{ The change of the fixed adaptive scale parameter $\tilde{s}_f$ and dynamic adaptive scale parameter $\tilde{s}_d^{(t)}$ when training on the cleaned WebFace dataset. The dynamic scale parameter $\tilde{s}_d^{(t)}$ gradually and automatically decreases to strengthen training supervisions for feature angles $\theta_{i,y_i}$, which validates our assumption on the adaptive scale parameter in our proposed AdaCos loss. Best viewed in color.
    }
\label{fig:s_AdaCos}

\end{figure}

\begin{figure}[t]
\begin{center}
   \includegraphics[width=1\linewidth]{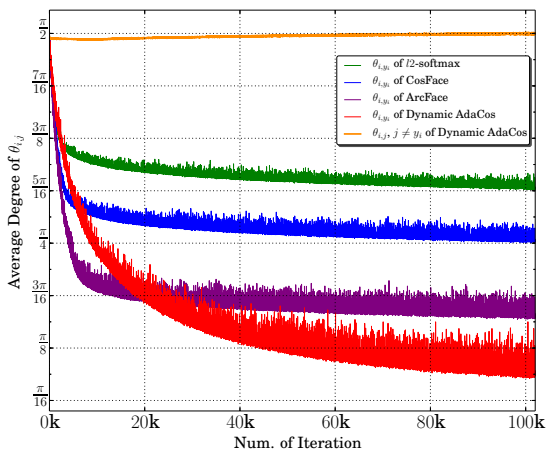}
\end{center}
   \caption{ 
         The change of $\theta_{i,y_i}$ when training on the cleaned WebFace dataset. $\theta_{i,y_i}$ represents the angle between the feature vector of $i$-th sample  and the weight vector of its ground truth  category $y_i$. Curves calculated by proposed dynamic AdaCos loss, $l2$-softmax loss \cite{L2-softmax}, CosFace \cite{CosFace} and ArcFace \cite{ArcFace} are shown. Best viewed in color.
    }
\label{fig:theta_AdaCos}

\end{figure}

Results in Table~\ref{tab:lfw_benchmark} show the recognition accuracies of models trained with different softmax loss functions. Our proposed AdaCos losses with fixed and dynamic scale parameters (denoted as Fixed AdaCos and Dyna. AdaCos) surpass the state-of-the-art cosine-based softmax losses under the same training configuration. For the hyperparameter settings of the compared losses, the scaling parameter is set as $30$ for $l2$-softmax~\cite{L2-softmax}, CosFace~\cite{CosFace} and ArcFace~\cite{ArcFace}; the margin parameters are set as $0.25$ and $0.5$ for CosFace~\cite{CosFace}, and ArcFace~\cite{ArcFace}, respectively. Since LFW is a relatively easy evaluation set, we train and test all losses for three times. The average accuracy of our proposed dynamic AdaCos is $0.26\%$ higher than state-of-the-art ArcFace~\cite{ArcFace} and $1.52\%$ than $l2$-softmax~\cite{L2-softmax}.

\subsubsection{Exploratory Experiments}
{\bf\quad The change of scale parameters and feature angles during training.}
In this part, we will show the change of scale parameter $\tilde{s}_d^{(t)}$ and feature angles $\theta_{i,j}$ during training with our proposed AdaCos loss. 
The scale parameter $\tilde{s}_d^{(t)}$ changes along with the current recognition performance of the model, which continuously strengthens the supervisions by gradually reducing $\theta_{i,y_i}$ and thus shrinking $\tilde{s}_d^{(t)}$.
Fig.~\ref{fig:s_AdaCos} shows the change of the scale parameter $s$ with our proposed fixed AdaCos and dynamic AdaCos losses. For the dynamic AdaCos loss, the scale parameter $\tilde{s}_d^{(t)}$ adaptively decreases as the training iterations increase, which indicates that the loss function provides stricter supervisions to update network parameters.
Fig.~\ref{fig:theta_AdaCos} illustrates the change of $\theta_{i,j}$ by our proposed dynamic AdaCos and $l2$-softmax. The average (orange curve) and median (green curve) of $\theta_{i,y_i}$, which indicating the angle between a sample and its ground-truth category, gradually reduce while the average (maroon curve) of $\theta_{i,j}$ where $j\neq{y_i}$ remains nearly $\frac{\pi}{2}$. Compared with $l2$-softmax loss, our proposed loss could achieve much smaller sample feature to category angles on the ground-truth classes and leads to higher recognition accuracies.

{\bf Convergence rates.}
Convergence rate is an important indicator of efficiency of loss functions. We examine the convergence rates of several cosine-based losses at different training iterations. The training configurations are same as Table~\ref{tab:lfw_benchmark}. Results in Table~\ref{tab:lfw_speed} reveal that the convergence rates when training with the AdaCos losses are much higher.

\begin{table}
\begin{center}
\begin{tabular}{|l|c|c|c|c|}
\hline
\multirow{2}{*}{Method} & \multicolumn{4}{c|}{Num. of Iteration}\\
\cline{2-5}& $25$k & $50$k & $75$k& $100$k \\
\hline\hline
Softmax                        & $70.15$ &$85.33$ &$89.50$ & $93.05$\\
$l2$-softmax~\cite{L2-softmax} & $79.08$ &$88.52$ &$93.38$ & $98.22$\\
CosFace~\cite{CosFace}         & $78.17$ &$90.87$ &$98.52$ & $99.37$\\
ArcFace~\cite{ArcFace}         & $82.43$ &$92.37$ &$98.78$ & $99.55$\\
\hline
Fixed AdaCos                 & $85.10$ &$94.38$ &$99.05$ & $99.63$\\
\bf{Dyna. AdaCos}                & $\textbf{88.52}$ &$\textbf{95.78}$ &$\textbf{99.30}$ & $\textbf{99.73}$\\
\hline
\end{tabular}
\end{center}
\caption{ Convergence rates of different softmax losses. At the same iterations, training with our proposed dynamic AdaCos loss leads to the best recognition accuracy.}
\label{tab:lfw_speed}

\end{table}

\subsection{Results on MegaFace}

\begin{table*}
\begin{center}
\begin{tabular}{|c||c|c|c|c|c|c|}

\hline
    \multirow{2}{*}{Method} & \multicolumn{6}{c|}{Size of MegaFace Distractor} \\
\cline{2-7}&$10^1$&$10^2$&$10^3$&$10^4$&$10^5$&$10^6$\\
    \hline  \hline
  $l2$-softmax& $99.73\%$ & $99.49\%$ & $99.03\%$ & $97.85\%$ & $95.56\%$ & $92.05\%$\\
    CosFace& $99.82\%$ & $99.68\%$ & $99.46\%$ & $98.57\%$ & $97.58\%$ & $95.50\%$\\  
    ArcFace   & $99.78\%$ & $99.65\%$ & $99.48\%$ & $98.87\%$ & $98.03\%$ & $96.88\%$\\
    \hline
 Fixed AdaCos& $99.85\%$ & $99.70\%$ & $99.47\%$ & $98.80\%$ & $97.92\%$ & $96.85\%$\\
    {\bf Dynamic AdaCos} & $\textbf{99.88\%}$ & $\textbf{99.72\%}$ & $\textbf{99.51\%}$ & $\textbf{99.02\%}$ & $\textbf{98.54\%}$ & $\textbf{97.41\%}$ \\
    
\hline

\end{tabular}
\end{center}
\caption{ Recognition accuracy on MegaFace by Inception-ResNet \cite{szegedy2017inception} models trained with different compared softmax loss and the same cleaned WebFace~\cite{WebFace} and MS1M~\cite{MS-Celeb-1M} training data.}
\label{tab:megaface_benchmark}
\end{table*}

We then evaluate the performance of proposed AdaCos on the MegaFace Challenge~\cite{MegaFace2}, which is a publicly available identification benchmark, widely used to test the performance of facial recognition algorithms.
The gallery set of MegaFace incorporates over $1$ million images from $690$K identities collected from Flickr photos~\cite{thomee2015yfcc100m}. We follow ArcFace \cite{ArcFace}'s testing protocol, which cleaned the dataset to make the results more reliable. We train the same Inception-ResNet~\cite{IR} models with CASIA-WebFace \cite{WebFace} and MS1M \cite{MS-Celeb-1M} training data, where overlapped subjects are removed. 

\begin{figure}[t]
\begin{center}
   \includegraphics[width=1\linewidth]{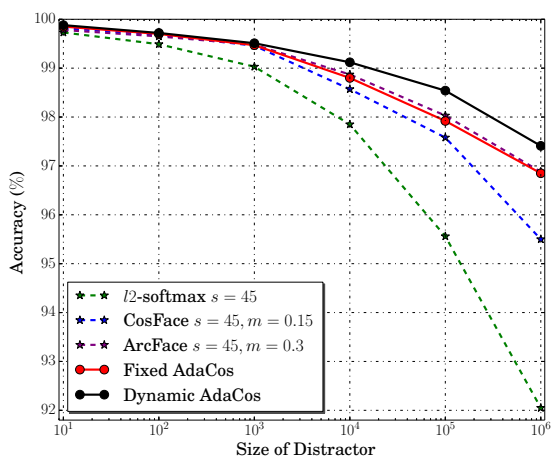}
\end{center}
   \caption{ Recognition accuracy curves on MegaFace dataset by Inception-ResNet \cite{szegedy2017inception} models trained with different softmax losses and on the same cleaned WebFace~\cite{WebFace} and MS1M~\cite{MS-Celeb-1M} training data. Best viewed in color.
    }
\label{fig:megaface}

\end{figure}

Table~\ref{tab:megaface_benchmark} and Fig.~\ref{fig:megaface} summarize the results of models trained on both WebFace and MS1M datasets and tested on the cleaned MegaFace dataset. The proposed AdaCos and state-of-the-art softmax losses are compared, where the dynamic AdaCos loss outperforms all compared losses on the MegaFace.

\subsection{Results on IJB-C 1:1 verification protocol}

\begin{table*}
\begin{center}
\begin{tabular}{|c||c|c|c|c|c|c|c|}

\hline
    \multirow{2}{*}{Method} & \multicolumn{7}{c|}{True Accept Rate @ False Accept Rate} \\
\cline{2-8}&$10^{-1}$&$10^{-2}$&$10^{-3}$&$10^{-4}$&$10^{-5}$&$10^{-6}$&$10^{-7}$\\
    \hline\hline  
FaceNet \cite{FaceNet}& $92.45\%$ & $81.71\%$ & $66.45\%$ & $48.69\%$ & $33.30\%$ & $20.95\%$&-\\

VGGFace \cite{parkhi2015deep}& $95.64\%$ & $87.13\%$ & $74.79\%$ & $59.75\%$ & $43.69\%$ & $32.20\%$&-\\

Crystal Loss \cite{ranjan2018crystal}& $99.06\%$ & $97.66\%$ & $95.63\%$ & $92.29\%$ & $87.35\%$ & $81.15\%$&$71.37\%$\\
    \hline
    
$l2$-softmax& $98.40\%$ & $96.45\%$ & $92.78\%$ & $86.33\%$ & $77.25\%$ & $62.61\%$ & $26.67\%$\\

CosFace \cite{CosFace}& $99.01\%$ & $97.55\%$ & $95.37\%$ & $91.82\%$ & $86.94\%$ & $76.25\%$ & $61.72\%$\\
    
ArcFace \cite{ArcFace}& $\textbf{99.07\%}$ & $\textbf{97.75\%}$ & $95.55\%$ & $92.13\%$ & $87.28\%$ & $82.15\%$ & $72.28\%$\\

    \hline
 Fixed AdaCos& $99.05\%$ & $97.70\%$ & $95.48\%$ & $92.35\%$ & $87.87\%$ & $82.38\%$ & $72.66\%$\\
    {\bf Dynamic AdaCos} & $99.06\%$ & $97.72\%$ & $\textbf{95.65\%}$ & $\textbf{92.40\%}$ & $\textbf{88.03\%}$ & $\textbf{83.28\%}$ & $\textbf{74.07\%}$ \\
    
\hline
\end{tabular}
\end{center}
\caption{ True accept rates by different compared softmax losses on the IJB-C 1:1 verification task. The same training data (WebFace \cite{WebFace} and MS1M \cite{MS-Celeb-1M}) and Inception-ResNet \cite{IR} networks are used. The results of FaceNet \cite{FaceNet}, VGGFace \cite{parkhi2015deep}, and Crystal Loss \cite{ranjan2018crystal} are from \cite{ranjan2018crystal}.}
\label{tab:ijbc11_benchmark}

\end{table*}
The IJB-C dataset \cite{ijbc} contains about $3,500$ identities with a total of $31,334$ still facial images and $117,542$ unconstrained video frames. In the 1:1 verification, there are $19,557$ positive matches and $15,638,932$ negative matches, which allow us to evaluate TARs at various FARs (\eg, $10^{-7}$).

\begin{figure}[t]
\begin{center}
   \includegraphics[width=1\linewidth]{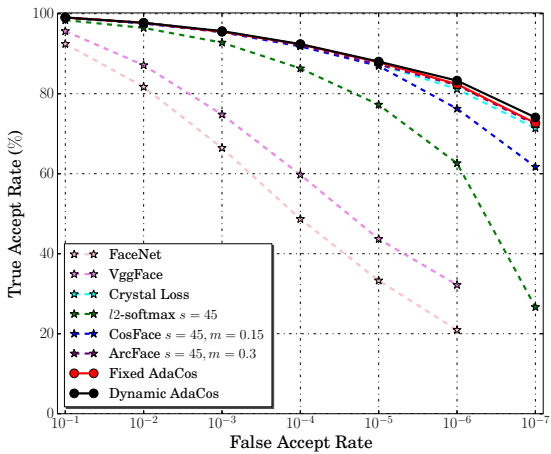}
\end{center}
   \caption{TARs by different compared softmax losses on the IJB-C 1:1 verification task. The same training data (WebFace \cite{WebFace} and MS1M \cite{MS-Celeb-1M}) and Inception-ResNet \cite{IR} are used. The results of FaceNet \cite{FaceNet}, VGGFace \cite{parkhi2015deep} are reported in Crystal Loss \cite{ranjan2018crystal}.}
\label{fig:ijbc11}

\end{figure}

We compare the softmax loss functoins, including the proposed AdaCos, $l2$-softmax \cite{L2-softmax}, CosFace \cite{CosFace}, and ArcFace \cite{ArcFace} with the same training data (WebFace \cite{WebFace} and MS1M \cite{MS-Celeb-1M}) and network architecture (Inception-ResNet \cite{IR}). We also report the results of FaceNet \cite{FaceNet}, VGGFace \cite{DeepFace} listed in Crystal loss~\cite{ranjan2018crystal}. Table \ref{tab:ijbc11_benchmark} and Fig.~\ref{fig:ijbc11} exhibit their performances on the IJB-C 1:1 verification. Our proposed dynamic AdaCos achieves the best performance.

\section{Conclusions}
In this work, we argue that the bottleneck of existing cosine-based softmax losses may primarily comes from the mis-match between cosine distance $\cos{\theta_{i,y_i}}$ and the classification probability $P_{i,y_i}$, which limits the final recognition performance. To address this issue, we first deeply analyze the effects of hyperparameters in cosine-based softmax losses from the perspective of probability. Based on these analysis, we propose the AdaCos which automatically adjusts an adaptive parameter $\tilde{s}_d^{(t)}$ in order to reformulate the mapping between cosine distance and classification probability. Our proposed AdaCos loss is simple yet effective. We demonstrate its effectiveness and efficiency by exploratory experiments and report its state-of-the-art performances on several public benchmarks.

\textbf{Acknowledgements.} This work is supported in part by SenseTime Group Limited, in part by the General Research Fund through the Research Grants Council of Hong Kong under Grants CUHK14202217, CUHK14203118, CUHK14205615, CUHK14207814, CUHK14213616, CUHK14208417, CUHK14239816, in part by CUHK Direct Grant, and in part by National Natural Science Foundation of China (61472410) and the Joint Lab of CAS-HK.

{\small
\bibliographystyle{ieee_fullname}
\bibliography{AdaCos}
}

\end{document}